\documentclass[10pt,twocolumn,letterpaper]{article}

\usepackage{iccv}
\usepackage{times}
\usepackage{epsfig}
\usepackage{graphicx}
\usepackage{subcaption}
\usepackage{bbm}
\usepackage{amsmath}
\usepackage{amssymb}

\usepackage{caption}
\usepackage[ruled,vlined]{algorithm2e}

\usepackage{multirow}
\usepackage{booktabs}
\usepackage{adjustbox}
\usepackage{threeparttable}
\newcommand{\tabincell}[2]{\begin{tabular}{@{}#1@{}}#2\end{tabular}}

\usepackage{textcomp}
\usepackage{wrapfig}
\usepackage{xcolor}
\usepackage[accsupp]{axessibility}  

\DeclareMathOperator*{\argmax}{arg\,max}

\usepackage[pagebackref=true,breaklinks=true,colorlinks,bookmarks=false]{hyperref}

\iccvfinalcopy 

\def\iccvPaperID{9419} 

\ificcvfinal\fi

\begin{document}

\title{ RANK-NOSH: Efficient Predictor-Based Architecture Search via \\ Non-Uniform Successive Halving }


\author{
{Ruochen Wang\textsuperscript{1},
\enskip Xiangning Chen\textsuperscript{1},
\enskip Minhao Cheng\textsuperscript{1},
\enskip Xiaocheng Tang\textsuperscript{2},
\enskip Cho-Jui Hsieh\textsuperscript{1}}\\
\textsuperscript{1}{Department of Computer Science, UCLA, \enskip \textsuperscript{2}DiDi AI Labs}\\
}

\maketitle
\ificcvfinal\thispagestyle{empty}\fi

\begin{abstract}
    Predictor-based algorithms have achieved remarkable performance in the Neural Architecture Search (NAS) tasks. However, these methods suffer from high computation costs, as training the performance predictor usually requires training and evaluating hundreds of architectures from scratch. Previous works along this line mainly focus on reducing the number of architectures required to fit the predictor. In this work, we tackle this challenge from a different perspective - improve search efficiency by cutting down the computation budget of architecture training. We propose NOn-uniform Successive Halving (NOSH), a hierarchical scheduling algorithm that terminates the training of underperforming architectures early to avoid wasting budget. To effectively leverage the non-uniform supervision signals produced by NOSH, we formulate predictor-based architecture search as learning to rank with pairwise comparisons. The resulting method - RANK-NOSH, reduces the search budget by $\sim5\times$ while achieving competitive or even better performance than previous state-of-the-art predictor-based methods on various spaces and datasets.
\end{abstract}

\section{Introduction}

Neural Architecture Search has demonstrated its effectiveness in discovering high-performance architectures for various computer vision tasks, including image classification \cite{sdarts, DARTS, pcdarts}, semantic segmentation \cite{autodeeplab, squeezenas}, and image generation \cite{ autogan-distiller, advnas, autogan}.
Concretely, NAS methods attempt to identify the best architecture from a vast search space according to a predefined performance metric (e.g., accuracy, latency, etc.).
Pioneering works in this field require training and evaluating thousands of architectures in the search space \cite{EvolvingDeep, nas, nasnet}.
For example, the reinforcement learning method proposed by Zoph \etal~\cite{nasnet} trains over 20,000 networks.
The tremendous amount of computation overhead largely limits their practical usage.
Since then, improving the efficiency of architecture search algorithms has become a central topic in the NAS community.

Recently, weight-sharing technique witnesses much success in improving the search efficiency of NAS \cite{oneshot, chen2021drnas, DARTS, enas, pcdarts}.
Those methods train a supernet that encompasses all architectures in the search space, and use the pretrained supernet to evaluate the performance of architectures.
Despite their search efficiency, weight-sharing methods are not generally applicable to arbitrary search spaces due to the restrictions in constructing supernets \cite{gates, 101}.
Moreover, they also suffer from various inductive biases caused by the weight-sharing mechanism \cite{sdarts, Shu2020Understanding, dartspt, rDARTS, weightsharing}, which has a tendency towards parameter-free operations and wide, shallow structures.

On the other hand, predictor-based NAS methods are free from the aforementioned disadvantages.
Starting from a pool of randomly selected architectures, previous methods iteratively conduct the following steps: 1) train and evaluate all the architectures in the pool fully; 2) fit a surrogate performance predictor; 3) use the predictor to propose new architectures and add them to the pool for the next round \cite{brpnas, bananas, arch2vec}.
Compared with previous RL and evolution-based NAS methods, using a performance predictor can reduce the number of networks evaluated from scratch.
However, training all the architectures in the candidate pool fully is still extremely computationally expensive.
Most complementary advances alone this line focus on developing better predictors that require a smaller training pool \cite{brpnas, bananas, arch2vec}, but the potential to further cut down the search cost by reducing the training length of individual architectures in the pool has not drawn much attention.

In this work, we aim to investigate the possibility of reducing the search cost of predictor-based NAS by reducing the number of epochs required to train every architecture in the candidate pool.
Inspired by successive halving \cite{sh}, our key idea is that the learning process of poor architectures can be terminated early to avoid wasting budgets.
However, it is non-trivial to integrate successive halving to predictor-based NAS formulations.
Firstly, predictor-based algorithms iteratively add new architectures to the candidate pool \cite{brpnas, bananas, arch2vec}, whereas regular successive halving only removes underperforming candidates from the initial pool.
Secondly, with successive halving, architectures in the pool will be trained for different number of epochs, so their validation accuracy are not directly comparable in a semantically meaningful way.
Standard regression-based predictor fitting, which requires the exact validation accuracy for each architecture when fully trained, will be problematic in this setting.


To tackle those challenges in a unified way, we propose RANK-NOSH, an efficient predictor-based framework with significantly improved search efficiency.
RANK-NOSH consists of two parts.
The first part is NOn-Uniform Successive Halving (NOSH), which describes a multi-level scheduling algorithm that allows adding new candidates and resuming terminated training process.
It is non-uniform in the sense that NOSH maintains a pyramid-like candidate pool of architectures trained for various epochs without discarding any candidates.
For the second part, we construct architecture pairs and use a pairwise ranking loss to train the performance predictor.
The predictor is essentially a ranking network and can efficiently distill useful information from our candidate pool consisting of architectures trained for different epochs.
Moreover, the proposed framework naturally integrates recently developed proxies that measure architecture performance without training \cite{zero-cost, nas_ntk, nas_wt}, which allows more architectures to be included in the candidate pool at no cost.

Extensive experimental evaluations on multiple search spaces, datasets, and budgets demonstrate the effectiveness and generality of the proposed method.
On DARTS space, NAS-Bench-101, and NAS-Bench-201, RANK-NOSH can reduce the search budget of SOTA predictor-based methods by 5x while achieving similar or even better results.

\section{Related Work}

\paragraph{One-shot NAS with Weight-Sharing}
One-shot NAS methods construct a weight-sharing supernet that encompasses all child models in the search space, and use the pretrained supernet to evaluate child models~\cite{oneshot, sdarts, chen2021drnas, DARTS, enas, pcdarts}.
In this paper, we separate one-shot NAS from predictor-based methods based on whether the weight-sharing technique is used.

\paragraph{Predictor-based NAS}
Predictor-based algorithms learn a surrogate performance predictor that can be used to propose new architectures \cite{gates, bananas, arch2vec}.
The surrogate predictor is defined as a regressor of the network's validation accuracy \cite{gates, bananas, arch2vec}.
Predictor-based methods have the advantage of being generally applicable to arbitrary DAG search spaces and free from the inductive biases caused by weight-sharing (e.g., bias towards parameter-free operations, wide and shallow structure)~\cite{sdarts, fairDARTS, Shu2020Understanding, dartspt, rDARTS, weightsharing}.
However, these methods also require full training of hundreds of architectures, which remains a major bottleneck of their search efficiency.
Existing works in this category mainly focus on improving the sample efficiency, i.e., reducing the number of architectures required to train the predictor \cite{gates, arch2vec}.
Their improvement mainly comes from a better architecture encoding, such as LSTM \cite{pnas, nao, alphax}, Path-Encoding \cite{bananas}, GCN \cite{brpnas, gates}, and unsupervised pretraining \cite{arch2vec}.



\paragraph{Learning to Rank}
The idea of adding pairwise comparisons to help training the predictors has been explored before~\cite{brpnas, gates, curve}.
Our work differs from them in the following two aspects:
1) Prior methods still require accurate validation accuracy obtained from fully-trained networks. Instead, we use pairwise ranking loss, which doesn't have such restriction and therefore can largely reduce the search cost.
2) Our motivation behind formulating the search problem as learning to rank is to effectively utilize non-uniform successive halving, but pairwise comparison mainly serves as a regularization in previous developments~\cite{brpnas, gates}.
Wistuba \etal~\cite{curve} proposes to study from a partial learning curve with extrapolation models.
Their work is orthogonal to ours, as their method can also be used to compare models at intermediate epochs in our Non-Uniform Successive Halving algorithm.

\paragraph{Successive Halving}
Successive halving~\cite{se} is an effective technique to reduce the search computation budget. 
It trains a pool of randomly generated configurations and gradually eliminates poor performers from the pool according to a predefined schedule.
It is adopted in Bandit literature \cite{se} and also studied in the context of hyperparameter optimization \cite{bohb, sh, hyperband}.
Liam \etal~\cite{rsws} also applies successive halving as a baseline.
Previous successive halving methods are uniform in the sense that candidates in the pool at any time are trained for the same number of epochs as it simply discards poor performers.
We extend successive halving to the non-uniform setting to support our iterative search algorithm.
In our method, new architectures are iteratively added to the pool, and we keep poor candidates to construct architecture pairs to perform predictor training.


\begin{figure}[t]
    \begin{center}
    \includegraphics[width=.9\linewidth]{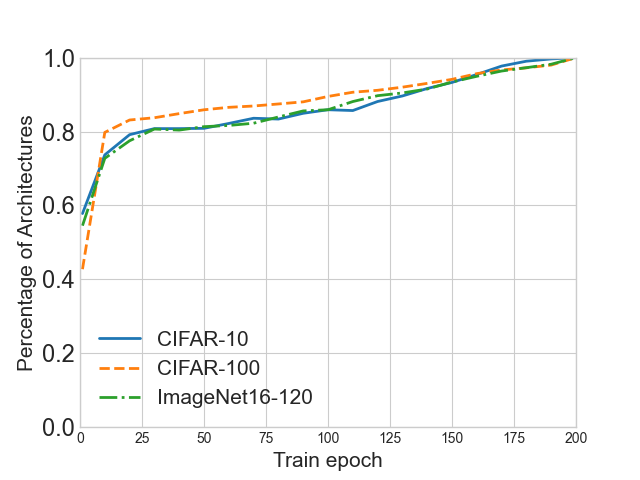}
    \end{center}
       \caption{Percentage of architectures with bottom-50\% validation accuracy at intermediate epochs that remain at bottom 50\% when fully trained on NAS-Bench-201.}
    \label{fig:corr-201-v2}
\end{figure}

\section{Methodology}
\label{sec:meth}

\begin{figure*}
\begin{center}
\includegraphics[width=1.\linewidth]{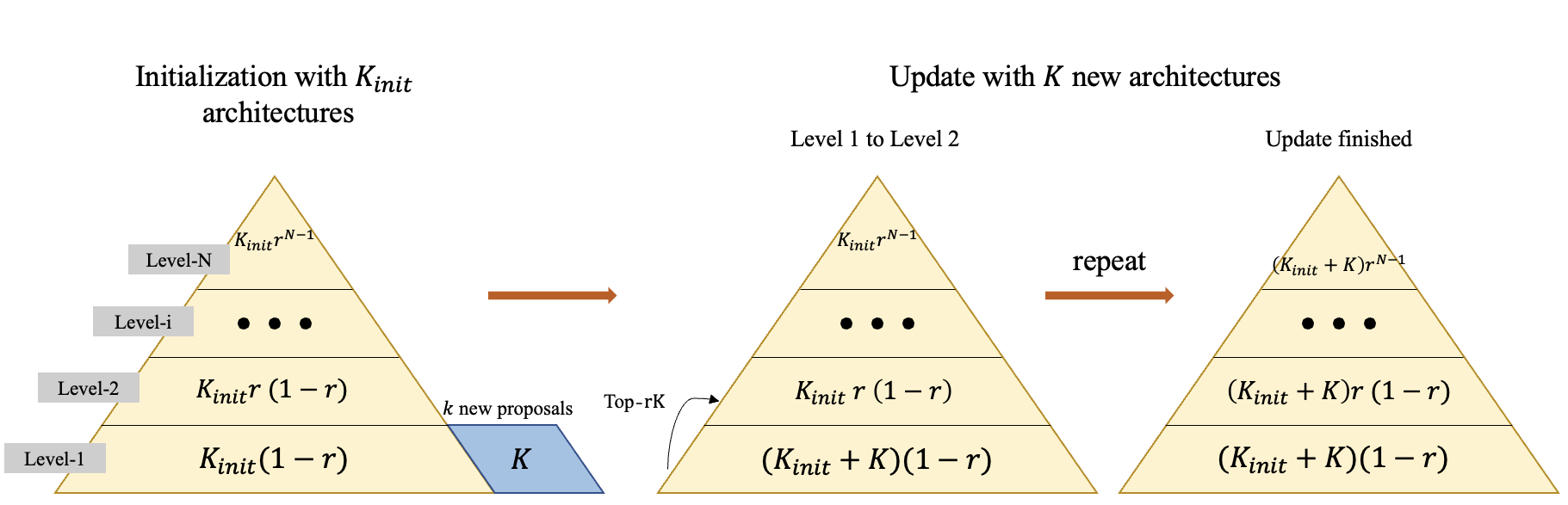}
\end{center}
   \caption{A N-level NOSH pyramid, including its initialization (left) and update (middle \& right) processes. Equation inside each level represents the corresponding number of architectures. All architectures in level-$i$ will be trained to epoch $e^{(i)}$.
   \textbf{Left:} During initialization, we populate the pool pyramid. Then we train the predictor and propose $K$ new architectures. \textbf{Middle:} We train the $K$ new candidates for $e^{(i)}$ epochs and move Top-$rK$ architectures from level-1 to level-2. \textbf{Right:} The pyramid after the update. Then we retrain the predictor and perform the next update, this process continues until a maximum pool size $M$ is achieved.}
\label{fig:pyramid}
\end{figure*}

In this section, we lay out the proposed RANK-NOSH framework.
We first introduce the motivations behind our method in Section \ref{sec:meth.motiv}.
The two key components of RANK-NOSH: Non-Uniform Successive Halving algorithm and search via learning to rank are described in Section \ref{sec:meth.nosh} and \ref{sec:meth.rank} respectively.
The complete search algorithm is provided in Section \ref{sec:meth.algo}.

\noindent {\bf Preliminaries} In this work, we focus on cell-based search space \cite{201, DARTS, 101} consisting of repeated searchable cells.
Each cell is represented as a Direct Acyclic Graph (DAG) $\mathcal{G} =  (\mathcal{V}, \mathcal{E})$, where $\mathcal{V}$ and $\mathcal{E}$ denote the set of nodes and edges respectively.
Each node in the DAG will be assigned an operation $o$ from the search space $|\mathcal{O}|$.
The discrete representations of these architectures can be characterized by the one-hot operation matrix $H \in \mathbb{R}^{|\mathcal{V}|*|\mathcal{O}|}$ and the adjacency matrix $A \in \mathbb{R}^{|\mathcal{V}|*|\mathcal{V}|}$, which will serve as the inputs for a GIN encoder~\cite{arch2vec}.
Note that the nodes here correspond to the edges in DARTS' DAG following Yan \etal~\cite{arch2vec}.

    \subsection{Motivation}
    \label{sec:meth.motiv}
        Existing predictor-based methods follow an iterative pipeline, which allows the predictor to focus on top performers and improves data efficiency \cite{brpnas}.
        Starting with a pool of randomly selected architectures, they train these architectures fully to obtain the validation accuracy as the regression label. Then they fit the performance predictor with those labels, and use it to propose new architectures that will be added to the pool for the next round.
        The computation cost of this pipeline is dominated by {\bf total number of epochs} required to train the architectures in the pool, which we refer to as the {\bf search budget}.
        For example, training a pool of 100 architectures to 200 epochs requires a search budget of 20,000 epochs, which significantly limits their practical usages.
        Therefore, reducing the search budget is crucial for speeding up  predictor-based NAS.
        
        During the search process of previous predictor-based methods, all architectures in the pool consume the same amount of training budget, regardless of their relative performance.
        Therefore, identifying underperforming architectures and terminating their training early could lead to significant savings.
        Figure~\ref{fig:corr-201-v2} shows that we can safely terminate inferior candidates at the early stage.
        For every training epoch, we plot the percentage of architectures with a bottom-50\% validation accuracy at the current epoch that remains in bottom-50\% when fully trained on NAS-Bench-201.
        As we can see, a majority of the poor architectures can be determined at early stages with increasing confidence as the training epoch increases.
        Specifically, when training for ten epochs, we observe that $\sim$ 70\% of architectures that lose at the starting line cannot catch up from behind when fully trained.
        Consequently, we can stop their training and spare the resources without a big sacrifice of the search performance.
        The trajectory of Spearman correlation in the Appendix further supports our observation.
        This is also the intuition behind early termination methods such as successive halving \cite{sh, hyperband} adopted in hyperparameter optimization.
        
        However, applying the idea of successive halving to predictor-based NAS requires special care.
        First, due to the iterative nature of predictor-based algorithms, the candidate pool keeps growing, while regular successive halving only removes candidates from the pool.
        Second, successive halving discards poor candidates at termination, resulting in a reduced number of training examples for the predictor.
        To solve the above issues in a unified framework, we propose RANK-NOSH that consists of two parts: a Non-Uniform Successive Halving (NOSH) scheduling algorithm that extends successive halving to handle growing candidate pools challenge, and a learning to rank algorithm to effectively utilize the non-uniform pool containing terminated candidates to train the performance predictor.
        Next, we discuss each part of the RANK-NOSH framework.

    \begin{algorithm}
        \SetAlgoLined
        \KwIn{Candidate pool $\mathcal{S}$, schedule $E = \{e^{(l)}\}_{l=1}^N$, move ratio $r$, Proposal size $K$ (use $K_{init}$ during the initialization round)}
        \KwResult{updated training pool $\mathcal{S}$}
        \For{level $l = 0 \sim (N - 1)$}{
            \eIf{$l == 0$}{
                Sort all architectures in level-$l$ according to their prior scores; \\
            }{
                Sort all architectures in level-$l$ according to their current validation accuracy; \\
            } 
            Train top $r K$ architectures in level-$l$ to epoch $e^{(l + 1)}$ and upgrade them to level-($l+1$); \\
            $K \mathrel{*}= r$;\\
        }
        \caption{NOSH: Non-Uniform Successive Halving}
        \label{algo:nosh}
    \end{algorithm}

    \subsection{Non-Uniform Successive Halving (NOSH)}
    \label{sec:meth.nosh}
        The key idea in NOSH is to maintain a pyramidal structure of the architecture pool that supports two operations: 1) \textit{Initialization}, which populates the pyramid with the initial pool, and 2) \textit{Update}, which inserts new candidates to the existing pyramid.
        We show a $N$-level NOSH Pyramid in Figure \ref{fig:pyramid}.
        Each level of the pyramid contains architectures trained for the same epoch (training epoch);
        and the training epoch increases as we level up, with the top level representing fully trained architectures.
        
        We introduce two parameters to control the level assignment - the training epoch schedule $E$ and move ratio $r$.
        The schedule $E = \{e^{(i)}\}_{i=1}^N$ represents the training epoch for every architecture at each level, where $e^{(i)} < e^{(i+1)}, \ i = 1 \sim (N - 1)$, and $e^{(N)}$ is the maximum number of epochs (fully trained).
        The move ratio $r \in (0,1)$ denotes the percentage of architectures moved when we proceeds from the current level to the next level.
        We then describe the process of initializing and updating the NOSH Pyramid below.
        
        \paragraph{Initialization}
        Starting with a pool of $K_{init}$ untrained architectures, we first train these architectures for $e^{(1)}$ epochs.
        From there, architectures with the validation accuracy in the bottom $K_{init}(1-r)$ will be terminated (kept in level 1), while the top $K_{init}r$ architectures will be trained further to $e^{(2)}$ epochs and upgrade to level 2. 
        This process repeats until the maximum training epoch $e^{(N)}$ is reached.
        After initialization, level-1 will contain $K_{init}(1-r)$ candidates and level-$N$ will contain $K_{init} {r}^{(N-1)}$ candidates.
        
        \paragraph{Update}
        Following previous predictor-based NAS algorithms \cite{brpnas, bananas, arch2vec}, we continuously propose new architectures and add them to the candidate pool. 
        In this case, the existing pyramid should be updated accordingly.
        After initialization, we train the predictor accordingly and propose new architectures, which will compete against the existing architectures in the pool to level up.
        Concretely, suppose there are $K$ new architectures (untrained) to be added to the pool.
        We first insert them into level-1 of the pyramid by training them for $e^{(1)}$ epochs.
        From there, the top $r K$ architectures will be selected to move to level-2, leaving $(K_{init}+K)(1-r)$ architectures in the new level-1 layer.
        The process repeats itself until we reach the top level.
        
        \paragraph{Train-free prior scores as level-0}
        Several recent works explore proxy metrics that produce a rough measurement of networks' performance without training,
        including metrics used for network pruning and Covariance of Jacobian at network initialization \cite{zero-cost, nas_ntk, nas_wt}.
        These metrics can be naturally integrated into our framework to allow more architectures added to the candidate pool at no cost.
        To do so, we simply add an extra level-0 to NOSH, where architectures are scored using training-free proxy metrics instead of their validation accuracy.
        These prior scores are cheap to evaluate, at the expense of low granularity especially among top configurations.
        Table~\ref{tab:spearman.201} illustrates the spearman correlation between architectures ranked by those training-free scores and validation accuracy when fully trained.
        As shown, the training-free metrics perform much better when ranking all architectures in the space than just the top-1\% architectures selected by their true validation accuracy. 
        Therefore, they can serve nicely as the level-0 information in our framework, since the distinction between top configurations can be refined at higher levels.
        
        We summarize NOSH process in Algorithm \ref{algo:nosh}.
        Note that NOSH differs from the regular successive halving in that 1) It is non-uniform in the sense that it maintains a candidate pool of architectures trained with different number of epochs without discarding any of them, which will be utilized to fit the ranker-based predictor. 2) Previously terminated architectures might have the chance to resume training if they outperform the newly added architectures.
        
        \begin{table}[!htb]
            \centering
            \caption{Spearman ranking correlation between architectures ranked by training-free metrics and true validation accuracy on CIFAR-10 in NAS-Bench-201 space.
            }
            \resizebox{.4\textwidth}{!}{
            \begin{threeparttable}
            \begin{tabular}{lcc}
            \hline

            {Prior Scores} & {Whole Space} & {Top 1\% Architectures} \\
            \hline
            grad\_norm \cite{zero-cost} & 0.58 & 0.42 \\
            jacob\_cov \cite{nas_wt} & 0.73 & 0.13 \\
            mag \cite{synflow} & 0.76 & 0.37 \\
            \hline
            \end{tabular}
            
            \end{threeparttable}}
            \label{tab:spearman.201}
        \end{table}

        \begin{figure}[t]
        \begin{center}
        \includegraphics[width=0.6\linewidth]{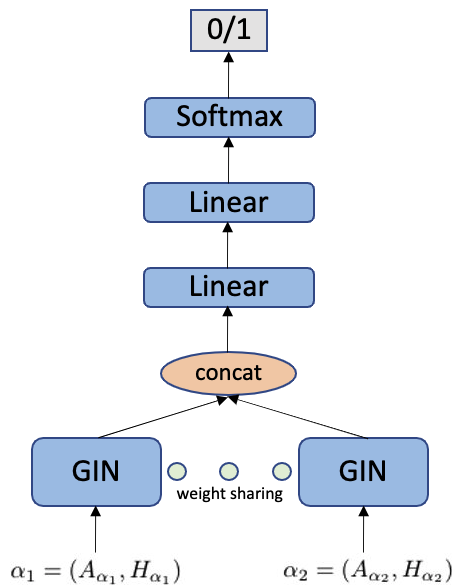}
        \end{center}
            \caption{Ranker Network}
        \label{fig:ranker}
        \end{figure}

    \subsection{NAS via Pairwise Ranking}
    \label{sec:meth.rank}
        After running NOSH, the candidate pool contains architectures trained with different number of epochs, where the pairwise ranking label of these architectures can be obtained directly: an architecture is considered to be better than another if it is either trained for more epochs (in a higher level of the pyramid), or have higher validation accuracy when their training epochs are the same (in the same level):
        \begin{align}
            \label{eq:comp}
            y(\alpha_1, \alpha_2) =
            \begin{cases} 
                \mathbbm{1}\{e_{\alpha_1} < e_{\alpha_2}\} & e_{\alpha_1} \neq e_{\alpha_2} \\
                \mathbbm{1}\{acc_{\alpha_1} < acc_{\alpha_2}\} & e_{\alpha_1} = e_{\alpha_2}, 
           \end{cases}
        \end{align}
        where $acc$ stands for the validation accuracy and $\mathbbm{1}$ is the indicator function.
        We therefore formulate the search process as learning to rank from pairwise $\{0,1\}$ labels, which naturally leverages the pairwise comparisons produced by NOSH.
        Concretely, given a pair of architectures, we use a ranker model to predict which one is better.
        The objective can be written as:
        \begin{align}
            \label{eq:obj}
            \min_{\mathcal{M}} \ & E_{(\alpha_1, \alpha_2) \sim \mathcal{X}} \big{[}\ell(\mathcal{M}(\alpha_1, \alpha_2), y(\alpha_1, \alpha_2))\big{]} \\
            \mathcal{X} &= \big{\{}(\alpha_1, \alpha_2) | \alpha_1 \in \mathcal{S}, \alpha_2 \in \mathcal{S}, \alpha_1 \neq \alpha_2\big{\}}, 
        \end{align}
        where $\mathcal{M}$ denotes the ranker model, $\ell$ is the loss function, and $\mathcal{X}$ denotes the set of all pairs of architectures from pool $\mathcal{S}$.
        At inference time, a global ranking among architectures of the whole space (or a large subspace) can be obtained to propose top architectures.
        
        We use a small Siamese network to model the ranker.
        The network consists of two MLPs on top of a pair of Siamese GIN encoders with shared weights.
        Figure \ref{fig:ranker} illustrates its structure.
        This ranker model is simple, yet expressive enough for our task, although using more advanced ranking networks \cite{deeprank} might further boost the performance.



        \begin{algorithm}
            \SetAlgoLined
            \KwIn{Max candidate pool size $M$, init pool size $K_{init}$, proposal size $K$, schedule $E = \{e^{(l)}\}_{l=1}^N$, move ratio $r$}
            \KwResult{Discovered best architecture $\alpha^{*}$}
            Randomly select $K_{init}$ architectures and add them to $\mathcal{S}$;\\
            Initialize Pyramid: $\mathcal{S} =$ NOSH($\mathcal{S}$, $E$, $r$, $K_{init}$);\\
            $M \mathrel{+}= K_{init}$; \\
            \While{$|\mathcal{S}| < M$}{
                Generate pairwise labels according to Eq.~(\ref{eq:comp}); \\
                Fit the ranker model with labeled $\mathcal{S}$;\\
                Use the ranker to propose top $min(K, M - |\mathcal{S}|)$ architectures and add them to $\mathcal{S}$;\\
                Update Pyramid: $\mathcal{S} =$ NOSH($\mathcal{S}$, $E$, $r$, $K$);\\
                $M \mathrel{+}= K$;
            }
            $\alpha^{*} = \argmax_{\alpha \in \mathcal{S}} Valid\_Acc_\alpha$
            \caption{RANK-NOSH Main Search}
            \label{algo:rank-nosh}
        \end{algorithm}

        \begin{table*}[!htb]
            \centering
            \captionsetup{justification=centering}
            \caption{Comparison with state-of-the-art NAS methods on NAS-Bench-201.}
            \resizebox{1.\textwidth}{!}{
            \begin{threeparttable}
            \begin{tabular}{lcccccccccc}
            \hline
            \multirow{2}*{\textbf{Method}} & \multicolumn{3}{c}{\textbf{CIFAR-10}} & \multicolumn{3}{c}{\textbf{CIFAR-100}} & \multicolumn{3}{c}{\textbf{ImageNet16-120}} \\ \cline{2-10}
            & validation & test & budget & validation & test & budget & validation & test & budget \\
            \hline
            
            DARTS~\cite{DARTS} & $39.77\pm 0.00$ & $54.30\pm 0.00$ & - & $38.57\pm 0.00$ & $38.97\pm 0.00$ & - & $18.87\pm 0.00$ & $18.41\pm 0.00$ & - & \\
            SNAS~\cite{snas} & $90.10\pm 1.04$ & $92.77\pm 0.83$ & - & $69.69\pm 2.39$ & $69.34\pm 1.98$ & - & $42.84\pm 1.79$ & $43.16\pm 2.64$ & - & \\
            GDAS~\cite{gdas} & $90.01\pm 0.46$ & $93.23\pm 0.23$ & - & $71.14\pm 0.27$ & $70.61\pm 0.26$ & - & $41.70\pm 1.26$ & $41.84\pm 0.90$ & - & \\
            PC-DARTS~\cite{pcdarts} & $89.96\pm 0.15$ & $93.41\pm 0.30$ & - & $67.12\pm 0.39$ & $67.48\pm 0.89$ & - & $40.83\pm 0.08$ & $41.31\pm 0.22$ & - & \\
            ENAS~\cite{enas} & $39.77\pm 0.00$ & $54.30\pm 0.00$ & - & $15.03\pm 0.00$ & $15.61\pm 0.00$ & - & $16.43\pm 0.00$ & $16.32\pm 0.00$ & - & \\
            \hline
            Prior Score: jacob\_cov~\cite{nas_wt} & $89.69\pm 0.73$ & $92.96\pm 0.80$ & - & $69.87\pm 1.22$ & $70.03\pm 1.16$ & - & $43.99\pm 2.05$ & $44.43\pm 2.07$ & - & \\
            Prior Score: mag~\cite{synflow} & $89.94\pm0.34$ & $93.35\pm0.04$ & - & $70.18\pm0.66$ & $70.47\pm0.18$ & - & $42.57\pm2.14$ & $43.17\pm2.57$ & - & \\
            \hline
            RE~\cite{re} \tnote{$\star$} & $91.04\pm0.51$ & $93.81\pm0.46$ & $1,200$ & $72.18\pm0.91$ & $72.06\pm0.97$ & $20,000$ & $45.78\pm0.72$ & $45.67\pm0.83$ & $20,000$ & \\
            RS~\cite{rs} \tnote{$\star$} & $90.91\pm0.41$ & $93.69\pm0.42$ & $1,200$ & $71.36\pm0.84$ & $71.32\pm0.95$ & $20,000$ & $45.26\pm0.67$ & $45.24\pm0.84$ & $20,000$ & \\
            REINFORCE~\cite{reinforce} \tnote{$\star$} & $90.32\pm0.85$ & $93.21\pm0.76$ & $1,200$ & $70.95\pm1.22$ & $70.87\pm1.23$ & $20,000$ & $44.66\pm1.44$ & $44.63\pm1.52$ & $20,000$ & \\
            arch2vec-BO~\cite{arch2vec} \tnote{$\star$} & $91.4\pm 0.35$ & $94.24\pm0.21$ & $1,200$ & $73.29\pm0.41$ & $73.41\pm0.22$ & $20,000$ & $46.27\pm0.39$ & $46.32\pm0.27$ & $20,000$ \\
            \hline
            RANK-NOSH & $\bf{91.4\pm0.18}$ & $\bf{94.26\pm0.17}$ & $\bf{292}$ & $\bf{73.49\pm0.00}$ & $\bf{73.51\pm0.00}$ & $\bf{5,550}$ & $\bf{46.37\pm0.0}$ & $\bf{46.34\pm0.0}$ & $\bf{5,550}$ \\
            \hline
            \textbf{oracle} & 91.61 & 94.37 & - & 73.49 & 73.51 & - & 46.77 & 47.31 & - \\
            \hline
            \end{tabular}
            
            \begin{tablenotes}
                \item[$\star$] Reproduced by directly searching on every dataset with a candidate pool size of 100 architectures following \cite{arch2vec}. Note that the original arch2vec paper \cite{arch2vec} measures the search budget in seconds, which translates to approximately 100 architectures on all three datasets.
            \end{tablenotes}
            \end{threeparttable}}
            \label{tab:201}
        \end{table*}

    \subsection{Search Algorithm}
    \label{sec:meth.algo}
        The search process is conducted in the standard iterative manner.
        First, we initialize the pool by randomly sampling $K_{init}$ architectures from the search space.
        The architectures in the pool will be updated via NOSH algorithm described above.
        After that, $\{0,1\}$ pairwise labels can be obtained for all pairs of architectures in the pool using Eq.~(\ref{eq:comp}), which will be used to fit the ranker model.
        Then, the ranker model sorts architectures in the search space, and proposes $K$ new architectures to be added to the pool for the next iteration.
        Since enumerating the entire search space is often expensive, we follow previous works \cite{arch2vec, brpnas} to use a large randomly selected subset of the search space instead.
        The above process is repeated until a predefined maximum candidate pool size $M$ is reached.
        Algorithm \ref{algo:rank-nosh} summarizes the entire search procedure.

\section{Experimental Results}
\label{sec:exp}
In this section we present empirical evaluations of the proposed method on three widely used search spaces: NAS-Bench-101, NAS-Bench-201 and DARTS space.
We compare the proposed method with previous SOTA predictor-based algorithms based on two metrics: 1) the best and average test errors of the searched architectures and 2) the {\bf search budget}, defined in Section \ref{sec:meth.motiv} as the {\bf total number of epochs} required to train all architectures in the pool.

    \subsection{Implementation Details}
    \label{sec:exp.impl}

        \paragraph{Ranker}
        We use arch2vec \cite{arch2vec} to pretrain the GIN encoder in the ranker model as it improves the quality of architectural representations.
        The detailed optimization and hyperparameter settings can be found in the Appendix.

        \paragraph{NOSH}
        For level $1\sim N$ that use the validation accuracy to evaluate architectures, we set $r$ to $\frac{1}{2}$.
        For level-0 that uses train-free prior scores, we reduce the ratio to $\frac{1}{3}$ to accommodate more architectures in the training pool at no cost. In all our experiments, the prior score is set as the magnitude of weights at initialization (``mag'' in Table \ref{tab:spearman.201}).
        For each search space, we determine the schedule $E$ according to the standard full training epoch for a fair comparison (ablate in Section~\ref{sec:ablate_sche}).
        Regarding the candidate pool size, we set the maximum pool size $M$ to the amount that leads to $\sim$ 5x speedup compared with previous predictor-based methods.
        The initial pool size $K_{init}$ is always set as $16*3$, and the architecture proposal size $K$ is set as $10*3$.
        Note that only $\frac{1}{3}$ of those architectures will consume search budget, as the rest of them remain untrained at level-0.
    
        \begin{table*}[!htb]
            \centering
            \captionsetup{justification=centering}
            \caption{Comparison with state-of-the-art NAS methods on DARTS Space.}
            \resizebox{.8\textwidth}{!}{
            \begin{threeparttable}
            \begin{tabular}{lcccccc}
            \hline
            
            \multirow{2}*{\textbf{Architecture}} &
            \multicolumn{2}{c}{\textbf{Test Error(\%)}} &
            \multirow{2}*{\textbf{\tabincell{c}{Param\\(M)}}}  &
            \multirow{2}*{\textbf{\tabincell{c}{Search Budget\\(\#epochs)}}}  &
            \multirow{2}*{\textbf{\tabincell{c}{Search\\Method}}}  &
            \\ \cline{2-3} & Best & Avg \\
            \hline
            RSWS~\cite{rsws} & $2.71$ & $2.85\pm 0.08$ & 4.3 & - & Weight Sharing \\
            DARTS~\cite{DARTS} & $2.76\pm0.09$\tnote{$\star$} & - & 3.6 & - & Weight Sharing \\
            SNAS~\cite{snas} & - & $2.85\pm 0.02$ & 2.8 & - & Weight Sharing \\
            BayesNAS~\cite{BayesNAS} & $2.81\pm 0.04$\tnote{$\star$} & - & 3.4 & - & Weight Sharing \\
            ProxylessNAS~\cite{proxylessnas} & \bf{2.08}\tnote{$\dagger$} & - & 4.0 & - & Weight Sharing \\
            ENAS~\cite{enas} & $2.89$\tnote{$\dagger$} & - & 4.6 & - & Weight Sharing \\
            P-DARTS~\cite{pDARTS} & $2.50$ & - & 3.4 & - & Weight Sharing \\ 
            PC-DARTS~\cite{pcdarts} & $2.57\pm0.07$\tnote{$\star$} & - & 3.6 & - & Weight Sharing \\
            SDARTS-ADV~\cite{sdarts} & - & $2.61\pm0.02$ & 3.3 & - & Weight Sharing \\
            \hline
            Random Search~\cite{DARTS} & $3.29\pm 0.15$\tnote{$\star$} & - & 3.2 & 2,400 & Random \\
            \hline
            GATES~\cite{gates} & $2.58$\tnote{$\dagger$} & - & 4.1 & 64,000 & Predictor \\
            BRP-NAS (high)~\cite{brpnas} & - & $2.59\pm0.11$ & - & 36,000 & Predictor \\
            BRP-NAS (med)~\cite{brpnas} & - & $2.66\pm0.09$ & - & 18,000 & Predictor \\
            BANANAS~\cite{bananas} & $2.57$ & $2.64$ & 3.6 & 5,000 & Predictor \\
            arch2vec-BO~\cite{arch2vec} & $\bf{2.48}$ & $2.56\pm 0.05$ & 3.6 & 5,000 & Predictor \\
            \hline
            RANK-NOSH & $2.50$ & $\bf{2.53\pm0.02}$ & 3.5 & $\bf{990}$ & Predictor \\
            \hline
            \end{tabular}
            
            \begin{tablenotes}
                \item[$\dagger$] Obtained on different search spaces than DARTS.
                \item[$\star$] Error bars are computed by retraining the best discovered architecture multiple times.
            \end{tablenotes}
            \end{threeparttable}}
            \label{tab:DARTS.cifar10}
        \end{table*}
    
    \begin{table}[!htb]
        \centering
        \caption{Comparison with SOTA methods on NAS-Bench-101. We report the avg test accuracy for our method over 10 random seeds.}
        \resizebox{.47\textwidth}{!}{
        \begin{tabular}{lcc}
        \hline
        
        {\textbf{Methods}} & {\textbf {\tabincell{c}{Search Budget \\(\#epochs)}}} & {\textbf{\tabincell{c}{Test Accuracy\\(\%)}}} \\
        \hline
        Prior Score: jacob\_conv \cite{nas_wt} & - & 89.11 \\
        Prior Score: mag \cite{synflow} & - & 92.66 \\
        \hline
        Random Search \cite{101} & 108,000 & 93.54 \\
        REINFORCE \cite{101} & 108,000 & 93.58 \\
        Regularized Evolution \cite{101} & 108,000 & 93.72 \\
        NAO \cite{nao} & 108,000 & 93.74 \\
        BANANAS \cite{bananas} & 54,000 & 94.08 \\
        arch2vec-BO \cite{arch2vec} & 43,200 & 94.05 \\
        \hline
        RANK-NOSH & 8,400 & 93.97 \\
        \hline
        
        \end{tabular}}
        \label{tab:101}
    \end{table}

    \subsection{Results on NAS-Bench-201}
    \label{sec:exp.201}
        NAS-Bench-201 \cite{201} is a recently developed search space that supports weight-sharing NAS methods.
        This benchmark contains 15,625 architectures evaluated on three datasets: CIFAR-10, CIFAR-100, and ImageNet16-120.
        Following previous works~\cite{201, arch2vec}, we use the results when training the architecture for 12 epochs on CIFAR-10, and 200 epochs on CIFAR-100 and ImageNet16-120.
        Therefore, we set $E = (1,2,3,12)$ for CIFAR-10, $E = (10,50,100,200)$ for CIFAR-100 and ImageNet16-120 to match the maximum training epochs.
        The candidate pool size $M$ is set as $100*3$, which amounts to a search budget of 292 epochs for CIFAR-10 and 5,550 epochs for CIFAR-100 and ImageNet16-120.
        
        As shown in Table \ref{tab:201}, RANK-NOSH obtains near-oracle performance on all three datasets, consistently outperforming previous predictor-based NAS methods with a search cost of only $24\%$ on CIFAR-10 and $28\%$ on CIFAR-100 and ImageNet16-120.

    \subsection{Results on NAS-Bench-101}
    \label{sec:exp.101}
        NAS-Bench-101 \cite{101} is a cell-based search space that provides validation and test accuracy of 423,624 architectures trained for 108 epochs on CIFAR-10.
        The search space is general, as each cell can have an arbitrary DAG structure that consists of at most seven nodes and nine edges.
        Since one-shot NAS with weight-sharing cannot be applied to this space \cite{101}, we compare RANK-NOSH with methods without weight-sharing exclusively.
        As NAS-Bench-101 only provides intermediate results for epoch 4, 12, 36, and 108, 
        we set the schedule $E = (12, 36, 108)$.
        The maximum candidate pool size $M$ is set to $200*3$, which amounts to a search budget of 8,400 total epochs.
        As shown in Table \ref{tab:101}, our method achieves competitive results than previous SOTA methods with $19\%$ of the budget.

    \subsection{Results on DARTS Space}
    \label{sec:exp.DARTS}
        The DARTS space \cite{DARTS} is the most widely used search space for evaluating NAS algorithms at scale.
        It contains two types of searchable cells: the normal cell that preserves spatial dimensions, and the reduction cell that halves the dimension.
        Following previous works~\cite{pnas, arch2vec}, we use the same cell structure for both normal and reduction cell, which amounts to $10^9$ possible architectures.
        This space is too large for predictor-based methods to enumerate, we therefore randomly sample 600k architectures from the full space and run our algorithm on this subset as did in previous works~\cite{brpnas, arch2vec}.

        \paragraph{CIFAR-10}
        We use a schedule of $(10,20,30,50)$ to match the maximum training epoch 50 used in previous predictor-based methods~\cite{bananas, arch2vec}.
        We set the maximum size of the candidate pool as $50*3$ architectures for this space.
        The resulting search budget is 990 epochs, which is only 1.65x the cost to retrain an architecture following the DARTS protocol (600 epochs) \cite{DARTS}.
        As a comparison, previous SOTA predictor-based methods like BANANAS~\cite{bananas} and arch2vec~\cite{arch2vec} use a search budget of 5000 epochs, which is 8.3x the cost of standard retraining.
        As a result, RANK-NOSH drastically improves search efficiency.
        
        We repeat the search algorithm under different random seeds and report the best and mean test errors of the architectures discovered.
        For architecture evaluation on CIFAR-10, we keep all the retrain settings identical to DARTS \cite{DARTS}.
        As shown in Table \ref{tab:DARTS.cifar10}, RANK-NOSH achieves a best test error of 2.50\% and an average test error of 2.53\% on CIFAR-10 with over 5x search budget reduction than previous SOTA predictor-based NAS methods.
        Our algorithm has better average performance and lower variance.
        
        \paragraph{ImageNet}
        We further evaluate the discovered architecture on ImageNet under transfer learning settings \cite{DARTS, arch2vec}.
        Table \ref{tab:DARTS.imagenet} shows that the discovered architecture achieves $25.2\%$ top-1 and $7.7\%$ top-5 test error, ranking top among NAS methods with comparable search spaces.
        
        \begin{table}[!htb]
            \centering
            \caption{Transfer learning results on ImageNet}
            \resizebox{.4\textwidth}{!}{
            \begin{threeparttable}
            \begin{tabular}{lcc}
            \hline
            
            {\textbf{Architecture}} & {\textbf{Test Error(\%)}} & {\textbf{Params (M)}} \\
            \hline
            NASNet-A~\cite{nasnet} \tnote{$\star$} & 26.0 & 5.3 \\
            AmoebaNet-A~\cite{amoebanet} \tnote{$\star$} & 25.5 & 5.1 \\
            PNAS~\cite{pnas} \tnote{$\star$} & 25.8 & 5.1 \\
            SNAS~\cite{snas} \tnote{$\star$} & 27.3 & 4.3 \\
            DARTS~\cite{DARTS} \tnote{$\star$} & 26.7 & 4.7 \\
            SDARTS-ADV~\cite{sdarts} & 25.2 & 4.8 \\
            arch2vec-BO~\cite{arch2vec} \tnote{$\star$} & 25.5 & 5.2 \\
            \hline
            RANK-NOSH & 25.2 & 5.3 \\ \hline
            \end{tabular}
            
            \begin{tablenotes}
                \item[$\star$] Results obtained from the arch2vec paper \cite{arch2vec}.
            \end{tablenotes}
            \end{threeparttable}}
            \vspace{-10pt}
            \label{tab:DARTS.imagenet}
    \end{table}

\section{Ablation Study}
\label{sec:ablation}
In this section, we conduct ablation studies on the proposed method.
We focus on evaluating search algorithms based on the {\bf validation accuracy}, which is directly optimized by predictor-based NAS methods.
NAS-Bench-201 is utilized in this section since it provides per-epoch results for all architectures in the space.
We use the 200-epoch version for all three datasets in NAS-Bench-201 in this section.
    
    \subsection{Train-free Prior scores}
    Table \ref{tab:101} and Table \ref{tab:201} include the results of solely based on training-free metrics for architecture search on NAS-Bench-101 and NAS-Bench-201.
    Following Mellor \etal~\cite{nas_wt}, we sample 1,000 architectures from the search space and then select the best architecture according to the training-free prior scores.
    It could be clearly observed that relying on prior scores alone leads to poor performance (worse than random search).
    The reason is that those scores cannot distinguish between top architectures as suggested by Table~\ref{tab:spearman.201}.

    \subsection{Comparison with Early Stopping}
    A straightforward way to reduce the search budget of predictor-based NAS is by early stopping, i.e., simply terminate all architectures at an intermediate epoch.
    This simple strategy does not train any architecture to the end, suffering the gap between intermediate and final epochs.
    We compare the proposed methods with arch2vec~\cite{arch2vec} + early stopping under various budgets.
    We set the candidate pool size to $100$ for arch2vec following their paper \cite{arch2vec}, and compute the termination epoch based on the corresponding budgets.
    As summarized in Table \ref{tab:es}, the performance of arch2vec with early stopping drops drastically at low budgets, whereas our method stays  relatively stable. RANK-NOSH also enjoys much smaller variance.

    \begin{table}[!htb]
        \centering
        \caption{Validation accuracy (\%) of the final architectures obtained by RANK-NOSH v.s. arch2vec-BO with early stopping on NAS-Bench-201.
        }
        \resizebox{.46\textwidth}{!}{
        \begin{tabular}{l|c|c|c}
        \hline
        
        {\textbf{Dataset}} & {\textbf{Search Budget}} & {\textbf {arch2vec-BO}} & {\textbf{RANK-NOSH}} \\
        \hline
        \multirow{2}*{CIFAR-10}
        & 5,550 & $91.00\pm0.61$ & $91.60\pm0.02$ \\
        & 2,969 & $90.35\pm0.62$ & $91.56\pm0.07$ \\
        \hline
        \multirow{2}*{CIFAR-100}
        & 5,550 & $73.23\pm0.61$ & $73.49\pm0.00$ \\
        & 2,969 & $71.88\pm1.19$ & $73.44\pm0.09$ \\
        \hline
        \multirow{2}*{ImageNet16-120}
        & 5,550 & $46.08\pm0.75$ & $46.37\pm0.00$ \\
        & 2,969 & $45.10\pm1.07$ & $46.43\pm0.21$ \\
        \hline

        \end{tabular}}
        \label{tab:es}
    \end{table}

    \begin{table}[!thb]
        \centering
        \caption{Validation Accuracy of final architectures from RANK-NOSH on CIFAR-10 under various schedules and move ratios. 
        Our method is relatively stable across various $E$ and $r$.
        }
        \begin{subtable}[h]{.5\textwidth}
            \centering
            \resizebox{.88\textwidth}{!}{
            \begin{tabular}{c|c|c}
                \hline
                {$E$} & {\textbf{Search Budget}} & {\textbf{Valid Accuracy ($\%$)}} \\
                \hline
                (10,50,200) & 6,750 & $91.60\pm0.03$ \\
                (10,50,100,200) & 5,550 & $91.60\pm0.02$ \\
                (5,25,50,200) & 4,075 & $91.59\pm0.03$ \\
                (5,10,25,200) & 3,400 & $91.57\pm0.06$ \\
                \hline
            \end{tabular}}
            \caption{Under different $E$}
            \label{tab:ablate.nosh.e}
        \end{subtable}
        \hfill
        \vspace{+2mm}
        \begin{subtable}[h]{.5\textwidth}
            \centering
            \resizebox{.64\textwidth}{!}{
            \begin{tabular}{c|c|c}
                \hline
                {$r$} & {\textbf{Search Budget}} & {\textbf{Valid Accuracy ($\%$)}} \\
                \hline
                0.7 & 9,750 & $91.58\pm0.06$ \\
                0.6 & 7,400 & $91.59\pm0.06$ \\
                0.5 & 5,550 & $91.60\pm0.02$ \\
                0.4 & 4,100 & $91.58\pm0.08$ \\
                0.3 & 2,950 & $91.40\pm0.16$ \\
                \hline
            \end{tabular}}
            \caption{Under different $r$}
            \label{tab:ablate.nosh.r}
         \end{subtable}
         \vspace{-15pt}
         \label{tab:ablate.nosh}
    \end{table}
    
    \subsection{NOSH Schedules}
    \label{sec:ablate_sche}
    Here we ablate the effect of different NOSH schedules on the proposed method.
    As discussed before, there are two parameters that define the resource allocation in NOSH: $E$ and $r$.
    We keep all other settings identical to Section \ref{sec:exp.201} and only vary those two parameters.
    We start with testing RANK-NOSH under different $E$ while fixing $r = \frac{1}{2}$.
    The results are summarized in Table \ref{tab:ablate.nosh.e}.
    Our method performs stably under different schedules regardless of the number of levels and epoch intervals.
    
    Next, we fix $E$ to $(10,50,100,200)$ as in Section~\ref{sec:exp} and vary $r$ for level-1 to $N$.
    As shown in Table~\ref{tab:ablate.nosh.r}, RANK-NOSH is robust for a wide range of $r$.
    Results for CIFAR-100 and ImageNet16-120 show similar trends to CIFAR-10.
    We include them in the Appendix due to space limitations.
    In practice, we recommend using $r = 0.5$ for level 1 to N as done in successive halving \cite{sh} and vary $E$ to accommodate for different search budgets.

\section{Conclusions}
We present RANK-NOSH, an efficient predictor-based NAS algorithm that significantly reduces the computation overhead.
Concretely, we propose Non-Uniform Successive Halving (NOSH) - a scheduling algorithm that terminates underperforming architectures early to avoid wasting budget, and formulate the search process as learning to rank to harness the pairwise comparison labels generated.
Experimental results on multiple search spaces and datasets demonstrate the effectiveness of the proposed method.
RANK-NOSH achieves comparable or even better results with a significantly reduced search cost. Moreover, the proposed framework could be extended to other applications.
For instance, RANK-NOSH can be applied to hyperparameter optimization by concatenating the hyperparameters with the architecture embeddings, which we will explore in future work.

\subsection*{Acknowledgements}
\noindent This work is supported by NSF under IIS-1901527, IIS-2008173, IIS-2048280 and by Army
Research Laboratory under agreement number W911NF-20-2-0158.

{\small
\bibliographystyle{ieee_fullname}
\bibliography{egbib}
}

\clearpage

\section{Appendix}
\label{sec:app}

\subsection{Spearman Correlation on NAS-Bench-201}
\label{sec:app.meth}
\begin{figure}[!htb]
        \begin{center}
        \includegraphics[width=.9\linewidth]{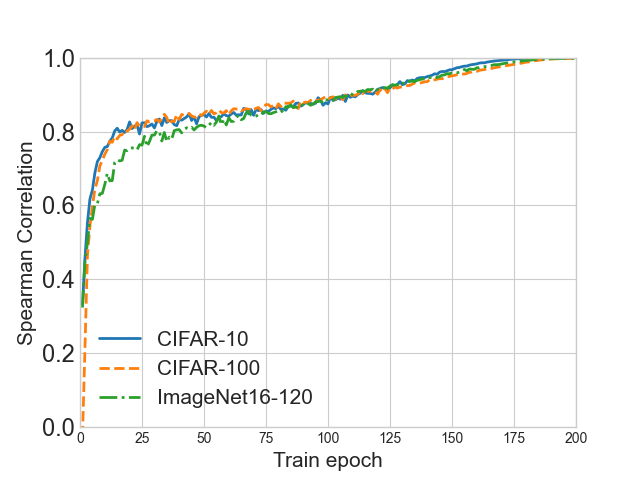}
        \end{center}
          \caption{Spearman ranking correlation between the validation accuracy of partially and fully trained architectures on NAS-Bench-201.}
        \label{fig:corr-201}
    \end{figure}

    We also plot the Spearman ranking correlation between the ranking of architectures at current epoch and that of fully trained ones for every training epoch.
    As showing in Figure \ref{fig:corr-201}, the ranking correlation reaches 0.6 only after a few epochs of training and increases steadily after that on all three datasets.
    The trajectory of ranking correlation serves as an extra piece of evidence that shows we can terminate the training of architectures at early stages to save the resources without a big sacrifice of the search performance.
    Moreover, due to the multi-level nature of the NOSH algorithm, each level obtains a more accurate ranking between architectures than previous levels.
    At the top level, architectures will be fully trained, leading to the true ranking among them in terms of the final validation accuracy.
    
\subsection{Search Spaces}
\label{sec:app.space}

    \paragraph{NAS-Bench-101}
    NAS-Bench-101~\cite{101} is a generic cell-based search space where the searchable operations are defined on the nodes in the cell, and the edges denote the data flow.
    Each searchable cell includes seven nodes, with the first being the input node and the last being the output node.
    There are four operations for each searchable node in this search space: {\tt conv\_1x1}, {\tt conv\_3x3}, {\tt conv\_5x5} and {\tt max\_pool\_3x3}, with {\tt conv\_5x5} approximated by two {\tt conv\_3x3}s.
    NAS-Bench-101 contains architectures with any arbitrary DAG structure between the input and output node with at most nine edges.
    
    \paragraph{NAS-Bench-201}
    The search cell in NAS-Bench-201 consists of five nodes (two input/output nodes and three intermediate nodes) and six edges.
    Unlike NAS-Bench-101, the operations are defined on edges in this space, and nodes represent data flow.
    Each edge is associated with one of the five operations: {\tt none}, {\tt skip}, {\tt conv\_1x1}, {\tt conv\_3x3}, and {\tt avg\_pool\_3x3}.
    Since we use the same GIN encoder as arch2vec~\cite{arch2vec}, we follow their method to transform this operation-on-the-edge representation into a graph where nodes present searchable operations and edges represent the data flow to match NAS-Bench-101.
    We refer the reader to the original arch2vec paper~\cite{arch2vec} for further details.
    
    The maximum training epoch on all three datasets is set to 200.
    However, for CIFAR-10, NAS-Bench-201 also provides a 12-epoch version, where architectures are trained for only 12 epochs with the learning rate scheduled accordingly.
    They use this version to bench its baselines on CIFAR-10.
    Therefore, for fair comparisons, we also use the 12-epoch version of CIFAR-10 for the main results in our experiment on NAS-Bench-201.
    
    \paragraph{The DARTS Space}
    Similar to NAS-Bench-201, the DARTS space ~\cite{DARTS} is an operation-on-the-edge space.
    Each cell contains six nodes, including two Input/Output nodes and four intermediate nodes.
    There are 14 possible edges in this search space, and eight of them will be selected to form an architecture.
    Every edge is associated with one of the following eight possible operations: {\tt none}, {\tt skip\_connect}, {\tt avg\_pool\_3x3}, {\tt max\_pool\_3x3}, {\tt sep\_conv\_3x3}, {\tt sep\_conv\_5x5}, {\tt dil\_conv\_3x3}, and {\tt dil\_conv\_5x5}.
    For the DARTS space, we also transform its DAG into the unified graph representation where nodes present searchable operations and edges represent the data flow as done in NAS-Bench-201~\cite{arch2vec}.

\subsection{Extended Discussion on Related Works}
In this work, we focus on task-agnostic methods for improving the efficiency of predictor-based NAS.
However, it is also possible to leverage task-specific prior knowledge to speed up architecture search.
For example, FCOS ~\cite{nasfcos} attempts to improve the search efficiency of RL-based NAS method on Object Detection tasks using fixed backbones and proxy tasks on VOC.
Since the techniques proposed in these works are task-dependent, we could not compare them with our method or previous SOTA NAS algorithms in the main experiments.

The concept of pausing and resuming the training of a candidate has been explored in Bayesian Optimization ~\cite{ftbo}:
FTBO ~\cite{ftbo} tries to decide when to pause or resume the training of a configuration via learning curve prediction.
In comparison, NOSH adopts a fixed schedule for simplicity.
In this sense, FTBO can be viewed as an orthogonal work to ours, and it might be an interesting future direction to study if FTBO could be applied to our framework to decide the NOSH schedules adaptively.
Moreover, FTBO focuses on the Hyperparameter Optimization task, whereas we mainly study Neural Architecture Search.

Neural Predictor~\cite{neuralpred} (NeuralPred) is an early and arguably the most straightforward predictor-based NAS algorithm.
It trains a neural network predictor on a pool of N fully trained architectures at once, and use it to propose K new architectures.
The total budget reported is $>100$ architectures, which is higher than the latest SOTA predictor-based methods we compared with, such as arch2vec-BO~\cite{arch2vec} and BANANAS~\cite{bananas}.
Moreover, NeuralPred focuses on the ProxylessNAS search space rather than the widely used DARTS Space.
For these reasons, we exclude the comparison with this method in the main experiments.
We encourage the readers to check out their paper for further details.

\subsection{Extra Details on the Experimental Settings}
\label{sec:app.exp}
    \paragraph{Ranker Network}
    As visualized in Figure \ref{fig:ranker}, the ranker network consists of two MLPs on top of a pair of Siamese five-layer GIN encoders with shared weights.
    The GIN encoder produces a 16-dimensional embedding for each architecture.
    And the feature embeddings from two architectures are concatenated into a 32-dim feature.
    The first MLP transforms this feature into a 64-dim hidden vector, which will then be mapped to a 2-dim output by the second MLP.
    The GIN encoders are pretrained using reconstruction loss following arch2vec~\cite{arch2vec};
    We refer the readers to their paper for further details of the pretraining step.
    
    We train the ranker network with a batch size of 10 for 100 epochs using Binary Cross-Entropy loss and Adam optimizer.
    The learning rate is set as 0.01 and annealed to 0.00001 with a cosine schedule.
    
    \paragraph{Train-free prior scores}
    Abdelfattah \etal~\cite{zero-cost} demonstrates that several metrics previously used for network pruning ~\cite{synflow} can serve as rough measures of architecture performance without training.
    In this work, we use the magnitude of model weights at initialization as our prior score due to its simplicity, although  more complex and advanced metrics can be deployed to further improve the performance.
    Concretely, after the network is initialized, we sum up the magnitude of its weights and use it as the score.
    The implementation is taken directly from the official Synaptic-Flow \cite{synflow} repo: \url{https://github.com/ganguli-lab/Synaptic-Flow}.
    
    \paragraph{Search algorithm}
    When proposing new architectures, we also deploy explicit exploration as done in BRP-NAS ~\cite{brpnas}.
    Concretely, the $K$ proposals are constructed by selecting the top $\frac{K}{2}$ architectures using global ranking and randomly sampling the rest half from the top $2K$ architectures (excluding top $\frac{K}{2}$ to avoid duplicates).
    This strategy allows RANK-NOSH to explore more diverse architectures in the search space.

\subsection{Complementary Results to Ablation Study}
\label{sec:app.ablation}

    \begin{table}[!thb]
        \caption{Validation Accuracy of final architectures from RANK-NOSH on CIFAR-10, CIFAR-100 and ImageNet16-120 under various schedules and move ratios. 
        Our method is relatively stable across various $E$ and $r$ on all three datasets.
        }
        \begin{subtable}[h]{.5\textwidth}
            \centering
            \resizebox{.9\textwidth}{!}{
            \begin{tabular}{l|c|c|c}
                \hline
                
                {\textbf Dataset} & {$E$} & {\textbf{Search Budget}} & {\textbf{Valid Accuracy ($\%$)}} \\
                \hline
                \multirow{4}*{CIFAR-10}
                & (10,50,200) & 6,750 & $91.60\pm0.03$ \\
                & (10,50,100,200) & 5,550 & $91.60\pm0.02$ \\
                & (5,25,50,200) & 4,075 & $91.59\pm0.03$ \\
                & (5,10,25,200) & 3,400 & $91.57\pm0.06$ \\
                \hline
                \multirow{4}*{CIFAR-100}
                & (10,50,200) & 6,750 & $73.49\pm0.00$ \\
                & (10,50,100,200) & 5,550 & $73.49\pm0.00$ \\
                & (5,25,50,200) & 4,075 & $73.42\pm0.22$ \\
                & (5,10,25,200) & 3,400 & $73.42\pm0.15$ \\
                \hline
                \multirow{4}*{ImageNet16-120}
                & (10,50,200) & 6,750 & $46.42\pm0.08$ \\
                & (10,50,100,200) & 5,550 & $46.37\pm0.00$ \\
                & (5,25,50,200) & 4,075 & $46.47\pm0.16$ \\
                & (5,10,25,200) & 3,400 & $46.33\pm0.27$ \\
                \hline
            \end{tabular}}
            \caption{Under different $E$}
            \label{tab:app.ablate.nosh.e}
        \end{subtable}
        \hfill
        \vspace{+2mm}
        \begin{subtable}[h]{.5\textwidth}
            \centering
            \resizebox{.7\textwidth}{!}{
            \begin{tabular}{l|c|c|c}
                \hline
                {\textbf Dataset} & {$r$} & {\textbf{Search Budget}} & {\textbf{Valid Accuracy ($\%$)}} \\
                \hline
                \multirow{4}*{CIFAR-10}
                & 0.7 & 9,750 & $91.58\pm0.06$ \\
                & 0.6 & 7,400 & $91.59\pm0.06$ \\
                & 0.5 & 5,550 & $91.60\pm0.02$ \\
                & 0.4 & 4,100 & $91.58\pm0.08$ \\
                & 0.3 & 2,950 & $91.40\pm0.16$ \\
                \hline
                \multirow{4}*{CIFAR-100}
                & 0.7 & 9,750 & $73.49\pm0.00$ \\
                & 0.6 & 7,400 & $73.46\pm0.11$ \\
                & 0.5 & 5,550 & $73.49\pm0.00$ \\
                & 0.4 & 4,100 & $73.46\pm0.11$ \\
                & 0.3 & 2,950 & $72.80\pm0.5$ \\
                \hline
                \multirow{4}*{ImageNet16-120}
                & 0.7 & 9,750 & $46.43\pm0.13$ \\
                & 0.6 & 7,400 & $46.50\pm0.25$ \\
                & 0.5 & 5,550 & $46.37\pm0.00$ \\
                & 0.4 & 4,100 & $46.40\pm0.08$ \\
                & 0.3 & 2,950 & $46.17\pm0.5$ \\
                \hline
            \end{tabular}}
            \caption{Under different $r$}
            \label{tab:app.ablate.nosh.r}
         \end{subtable}
         \vspace{-15pt}
         \label{tab:app.ablate.nosh}
    \end{table}

    We include extra ablation study results in this section.
    All experiments are conducted by running the search algorithm for 10 random seeds, as done in Section \ref{sec:ablation}.
    
    \paragraph{Results on other datasets}
    We provide ablation study results for NOSH schedules on all three datasets on NAS-Bench-201.
    As shown in Table \ref{tab:app.ablate.nosh}, RANK-NOSH is stable under a wide range of schedules and move ratios across datasets.

    \paragraph{Effectiveness of the ranker model in RANK-NOSH}
    To leverage the non-uniform signals produced by the NOSH algorithm, we adopt a ranker model as the performance predictor, optimized with discrete pairwise ranking loss.
    Compared with regression models, one potential downside of discrete pairwise loss is that it discards the fine-grain numerical values of validation accuracy.
    However, ranking-based methods also increase sample efficiency by creating $O(N^2)$ (pairs of) data points out of $N$ original samples, which may cancel out the loss of fine-grain information.
    Empirically, we observe a net gain of the adopted ranker model over the regression model used in arch2vec.
    To show this, we compare RANK with arch2vec at full budget (i.e., without early stopping or NOSH).
    As shown in Table \ref{tab:ranker_alone}, the ranker model alone leads to a near-oracle validation accuracy of $91.6\%/73.49\%/46.71\%$, outperforming arch2vec-BO.
    
    \begin{table}[!htb]
    \centering
        \caption{Validation accuracy (\%) of the final architectures obtained by RANK and arch2vec-BO at full budget on NAS-Bench-201.
        }
        \resizebox{.46\textwidth}{!}{
        \begin{tabular}{l|c|c|c}
        \hline
        
        {\textbf{Dataset}} & {\textbf{Search Budget}} & {\textbf {arch2vec-BO}} & {\textbf{RANK-NOSH}} \\
        \hline
        \multirow{1}*{CIFAR-10}
        & 20,000 (100\%) & $91.48\pm0.16$ & $91.60\pm0.02$ \\
        \hline
        \multirow{1}*{CIFAR-100}
        & 20,000 (100\%) & $73.29\pm0.41$ & $73.49\pm0.00$ \\
        \hline
        \multirow{1}*{ImageNet16-120}
        & 20,000 (100\%) & $46.27\pm0.39$ & $46.71\pm0.12$ \\
        \hline

        \end{tabular}}
        \label{tab:ranker_alone}
    \end{table}
    
    \paragraph{Effectiveness of the NOSH algorithm in RANK-NOSH}
    To further validate the necessity of NOSH algorithm over naive early stopping (ES), we compare RANK-NOSH with RANK-ES, i.e., replace the NOSH algorithm in the proposed method with early stopping.
    As shown in Table \ref{tab:rank_es}, RANK-NOSH consistently outperforms RANK-ES, demonstrating that the NOSH algorithm is critical to our framework.
    Note that from Table \ref{tab:es} and Table \ref{tab:rank_es}, we can see that the performance of ES is quite unstable over multiple runs;
    We conjecture that it is because the noisy signals produced by early stopping might mislead both predictor and final selection, resulting in much larger variances.
    
    \begin{table}[!htb]
        \centering
        \caption{Validation accuracy (\%) of the final architectures obtained by RANK-NOSH v.s. RANK-ES on NAS-Bench-201.
        }
        \resizebox{.46\textwidth}{!}{
        \begin{tabular}{l|c|c|c}
        \hline
        
        {\textbf{Dataset}} & {\textbf{Search Budget}} & {\textbf {RANK-ES}} & {\textbf{RANK-NOSH}} \\
        \hline
        \multirow{1}*{CIFAR-10}
        & 2,969 & $91.16\pm0.32$ & $91.56\pm0.07$ \\
        \hline
        \multirow{1}*{CIFAR-100}
        & 2,969 & $72.46\pm0.30$ & $73.44\pm0.09$ \\
        \hline
        \multirow{1}*{ImageNet16-120}
        & 2,969 & $45.42\pm1.01$ & $46.43\pm0.21$ \\
        \hline

        \end{tabular}}
        \label{tab:rank_es}
    \end{table}

\subsection{Discovered Architectures}
\label{sec:app.genotypes}
\begin{figure}[!htb]
        \begin{center}
        \includegraphics[width=1.\linewidth]{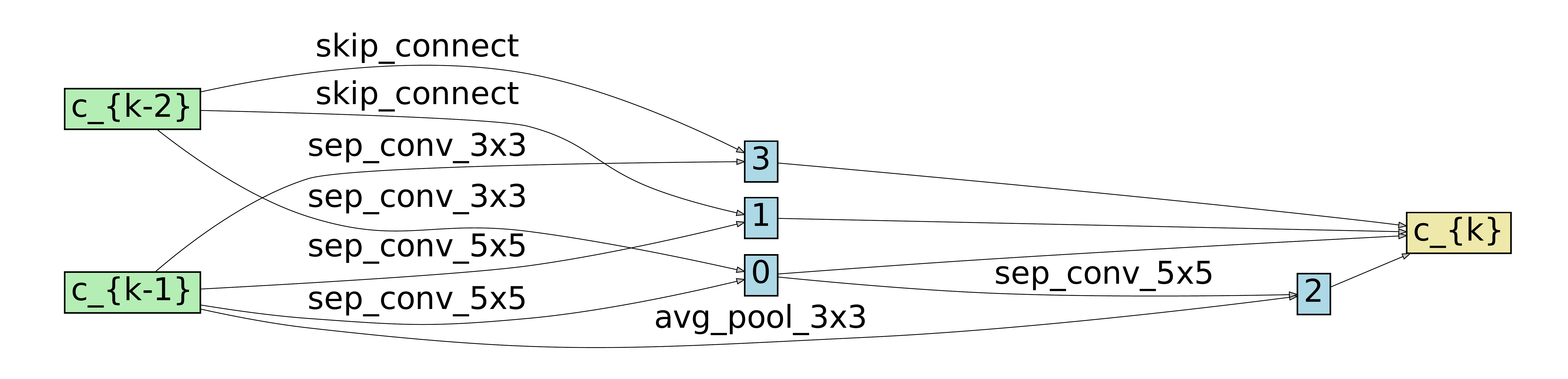}
        \end{center}
          \caption{Cell Discovered by RANK-NOSH on the DARTS space.}
        \label{fig:genotype}
    \end{figure}

    The best architecture discovered by RANK-NOSH on the DARTS space is visualized in Figure \ref{fig:genotype}.
    As mentioned in the main text, we follow arch2vec~\cite{arch2vec} and use the same cell for both reduction and normal cells.
    
\end{document}